\documentclass[11pt]{article}

\usepackage[preprint]{acl}
\usepackage{enumitem}
\usepackage{times}
\usepackage{latexsym}
\usepackage[T1]{fontenc}
\usepackage[utf8]{inputenc}
\usepackage{microtype}
\usepackage{inconsolata}
\usepackage{graphicx}
\usepackage{amsmath}
\usepackage{booktabs}
\usepackage{array}
\usepackage{xcolor}
\usepackage{pifont}
\newcommand{\corpusname}{\textsc{EduWild}}
\usepackage{tabularx}
\usepackage{siunitx}
\definecolor{cFactual}{HTML}{A64453}
\definecolor{cWorkflow}{HTML}{3E62AF}
\definecolor{cDiagnose}{HTML}{478957}
\definecolor{cPedagogical}{HTML}{B2781F}
\newcommand{\codeF}[1]{\textcolor{cFactual}{\textbf{#1}}}
\newcommand{\codeW}[1]{\textcolor{cWorkflow}{\textbf{#1}}}
\newcommand{\codeD}[1]{\textcolor{cDiagnose}{\textbf{#1}}}
\newcommand{\codeP}[1]{\textcolor{cPedagogical}{\textbf{#1}}}
\newcommand{\cbentry}[3]{\par\smallskip\noindent#1: \textbf{#2} is defined as #3}

\title{EduWild: How Teacher-Configured Chatbots Fail in Real-World Use and How to Fix Them}

\author{Ziqi Liu \quad Ruiwei Xiao\thanks{Corresponding authors.} \quad Yuxuan Li\footnotemark[1] \quad John Stamper \\
Carnegie Mellon University \\
{\small\normalfont\texttt{ziqi@cmu.edu}\quad\texttt{ruiweix@cs.cmu.edu}} \\
{\small\normalfont\texttt{yuxuanll@andrew.cmu.edu}\quad\texttt{jstamper@cmu.edu}}}

\begin{document}
\maketitle

\begin{abstract}
Teachers increasingly configure educational chatbots through natural-language instructions, but how these bots fail in real-world use remains poorly understood.
We introduce \corpusname, a dataset of 1,726 conversations with 217 teacher-configured chatbots, each paired with the instructions and configuration behind it, including 900 conversations from the intended end users of 35 deployed chatbots.
We develop a 29-code taxonomy covering factual, workflow, diagnostic, and pedagogical failures and apply it using a human-calibrated LLM judge.
The most frequent failures are lapses in following teacher intent --- bots skip required workflow steps (53.8\% of end-user conversations), omit expectation setup (33.4\%), and overload learners with information or questions (37.1\%).
Guided by these findings, we design a prompt-based harness targeting the six most frequent failure modes.
In a study with 200 college students across ten bots, the harness reduces the mean number of target failure codes per conversation from 1.406 to 0.489, a 65.2\% relative reduction with bots weighted equally.
These findings show how real-world failure analysis can guide practical improvements to educational chatbots.
\end{abstract}

\section{Introduction}
\label{sec:introduction}

Large language models (LLMs) enable chatbots to answer course questions, guide problem solving, and provide feedback \citep{macina2023mathdial,wang2024bridging}, with teachers configuring their behavior through natural-language instructions \citep{team2024learnlm}.
As these systems are deployed in educational settings \citep{bastani2025generative}, evaluating their reliability requires attention to instruction following and the quality of multi-turn interactions \citep{team2025evaluating,macina2025mathtutorbench}. Yet we have limited understanding of the failures that occur when teacher-created chatbots are used across different tasks.
Real-world conversation data are needed to identify these failures and improve chatbot behavior.

Existing datasets primarily capture human tutoring or interactions with simulated students \citep{macina2023mathdial,wang2024bridging,caines2022teacher}, while benchmarks evaluate pedagogical capabilities in controlled settings, often focusing on mathematics \citep{maurya2025unifying,macina2025mathtutorbench}.
These resources offer limited insight into how teacher-created chatbots fail in everyday use.

To address this gap, we collect and introduce \corpusname, a dataset of 1,726 real-world conversations with 217 teacher-configured educational chatbots on Playlab, covering tutoring and coaching, question answering, and task assistance.
Each conversation is paired with the teacher's system prompt and configuration, so chatbot behavior can be judged against what the teacher intended.
We develop a 29-code taxonomy to characterize factual, workflow, diagnostic, and pedagogical failures, drawing on research on scaffolding, adaptive instruction, and pedagogical evaluation \citep{macina2023mathdial,wang2024bridging,team2025evaluating}.
Applying the taxonomy to the 900 end-user conversations with a human-calibrated LLM judge shows that the most common failures concern following teacher intent.
Chatbots skip required workflow steps in 53.8\% of conversations, more than five times the rate of incorrect content (10.1\%).
They also fail to establish their role and scope, and overload users with information or multiple questions at once.

To address these failures, we develop a harness, a prompt-level wrapper around each teacher's original chatbot, that guides chatbots to follow task requirements, establish expectations, and keep responses and questions manageable.
We evaluate the harness in a Prolific study with 200 participants across ten chatbots.
Participants are randomly assigned to either a control condition using the original chatbot or an experimental condition using the chatbot with the harness, with 100 participants in each condition.
The harness reduces the average number of distinct target failure modes per conversation by 65.2\%, with the largest improvements in multi-question overload and cognitive overload.
These results show how failure analysis of real-world educational dialogues can guide targeted improvements in chatbot behavior.

Our contributions are:
\begin{itemize}
    \item \corpusname, to our knowledge the first dataset of real-world educational chatbot conversations paired with the teacher-authored instructions and configuration behind each chatbot.
    \item A 29-code failure taxonomy and an analysis of \corpusname, highlighting frequent failures to follow teacher-defined workflows and respect teacher-defined boundaries.
    \item A harness informed by the observed failures, evaluated through a human participant study.
\end{itemize}

\section{Related Work}
\label{sec:related-work}

\subsection{Educational Dialogue and Pedagogical Control}

Educational dialogue systems range from AutoTutor's mixed-initiative tutoring \citep{graesser2005autotutor} to LLM-based approaches such as LearnLM, which follows pedagogical system instructions \citep{team2024learnlm}, and Tutor CoPilot, which supports human tutors and improves student topic mastery \citep{wang2024tutor}.
Classroom evidence further shows that tutoring guardrails affect subsequent learning beyond assisted task performance \citep{bastani2025generative}.
Together, these studies motivate evaluating instructional behavior alongside answer quality.
We examine how deployed chatbots execute diverse teacher-authored roles and where that execution fails.

\subsection{Educational Conversation Data and Evaluation}

Educational dialogue resources include teacher--simulated-student conversations in MathDial \citep{macina2023mathdial}, real mathematics tutoring with expert remediation in Bridge \citep{wang2024bridging}, and authentic English tutoring annotated for conversational sequences or pedagogical strategies in TSCC v2 and TACT \citep{caines2022teacher,yang2026tact}.
MRBench evaluates responses to student mistakes across pedagogical dimensions \citep{maurya2025unifying}, while MathTutorBench distinguishes problem-solving from teaching ability and identifies challenges in longer dialogues \citep{macina2025mathtutorbench}.
Simulated students offer another evaluation route but differ substantially from real learners \citep{scarlatos2026simulated}.
Complementing these resources, \corpusname captures real-world conversations with teacher-configured chatbots across educational roles, supporting conversation-level failure analysis against teacher instructions and mitigation evaluated with human participants.

\section{\corpusname{} Dataset}

\subsection{Data Collection}

\paragraph{Source and context.}
We collected \corpusname{} through a self-paced, nine-week online professional development program on AI in education for higher-education educators, offered to 114 cohorts between February 2025 and July 2026.
Building at least one educational chatbot was a required learning activity in the program, and educators could voluntarily share their chatbots with intended end users: students, fellow educators and staff such as librarians, or the general public.
Chatbots were authored and deployed on Playlab\footnote{\url{https://www.playlab.ai}}, a nonprofit, no-code platform on which educators build chatbots by writing natural-language system prompts and share them through a link.

\paragraph{Collected data.}
For each chatbot, we export its system prompt and configuration (title, description, AI model, attached files, and platform settings).
For each conversation, we export the full transcript with speaker roles, per-message timestamps, and HTML formatting. In total, 87 educators created 224 chatbots, which logged 1,744 conversations.

\paragraph{Preprocessing.}
We exclude test activity by research team members (6 chatbots and 4 conversations), one chatbot without a clear educational purpose and its 8 conversations, and six conversations with no message content, leaving 217 chatbots by 85 educators and 1,726 conversations.
We consider a chatbot \emph{deployed} if its author published it and it has at least one conversation from an anonymous user accessing it through a shared link; 35 chatbots meet this criterion (28 for students, 6 for educators and staff, and 1 for a general audience).
We assign each conversation to one of three user groups: end users of the deployed chatbots (900 conversations), chatbot authors (741), and other educators in the program (85).
The dataset retains all three groups, but our failure analysis uses only end-user conversations, since author and peer conversations reflect testing and exploration rather than use by the intended audience.
To de-identify the data, we replaced personal information with placeholders in both transcripts and system prompts.

\paragraph{Ethics.}
The study was approved by our institution's IRB (protocol number withheld for review).
Before interacting with Playlab, both educators and end users agreed to a data use policy and provided informed consent for the research use of their chatbots and conversation logs.
End users accessed chatbots through anonymous links without accounts, so conversations carry no user identifiers.
All educators were adults; all end users were adults as all chatbots were deployed in higher-education courses and the data use policy required users to confirm they were 18 or older.

\paragraph{Release.} Upon publication, we will release the codebook, annotation prompts, and analysis code publicly, and make the de-identified dataset available to researchers under a data use agreement that prohibits re-identification and commercial use.

\subsection{Dataset Statistics}
\label{sec:stats}

\corpusname{} comprises two layers of data: \emph{authoring data}, the system prompts and configurations of 217 chatbots built by 85 educators, and \emph{interaction data}, 1,726 conversation transcripts with these chatbots.
Both layers come from authentic educational practice: educators built chatbots for their own courses or colleagues, and end users used them in practice.
Our analyses focus on the 35 deployed chatbots and their 900 end-user conversations.
Appendix~\ref{app:schema} lists the data schema exported for authoring and conversation data and corresponding interfaces.

\paragraph{Authoring data.}
Educators authored 217 chatbots, with a median of one and a maximum of 14 per educator (Table~\ref{tab:authoring}).
System prompts are moderately long ($\mathit{Mdn} = 296$ words), and those of deployed chatbots are substantially longer ($\mathit{Mdn} = 500$).
Beyond the prompt, 70\% of chatbots attach at least one reference file, such as a syllabus, rubric, or textbook, rising to 91\% among deployed chatbots.
Built-in platform features are used sparingly: 8.3\% enable the setting that discourages direct answers, 9.7\% accept image input, and 10.6\% enable any tool, with each rate 1.6--2.8 times higher among deployed chatbots.
At export, 73\% of chatbots ran on Claude models, nearly all Claude Sonnet 4.6 (91\%), and 23\% on GPT models, nearly all GPT-5.4 (98\%), rising to 86\% Claude among deployed chatbots.

We categorized each chatbot by intended users, function, and subject based on its title, description, and system prompt. One author drafted labels for all 217 chatbots with LLM assistance (GPT-6 Astra) and manually reviewed each label.
Most chatbots target students (59\%), spanning writing and language, arts and humanities, STEM, health sciences, and other disciplines.
The largest subject, however, is teaching and course design (30\%), almost entirely aimed at educators themselves for tasks such as drafting learning objectives, redesigning assessments, and creating accessible learning materials.
By function, supporting tasks is most common (45\%), followed by providing information (26\%) and facilitating learning (24\%); educator- and staff-facing chatbots nearly all support tasks, whereas student-facing chatbots mostly facilitate learning or provide information.
Deployed chatbots skew further toward students (80\%) and toward providing information (37\%), with writing and language the most common subject (31\%).

\begin{table}[t]
\centering
\footnotesize
\setlength{\tabcolsep}{4pt}
\begin{tabular}{lrr}
\toprule
 & \textbf{All} & \textbf{Deployed} \\
\midrule
\multicolumn{3}{@{}l}{\textit{Overview}} \\
\quad Chatbots & 217 & 35 \\
\quad Educators & 85 & 15 \\
\quad Chatbots per educator$^\dagger$ & 1 & 2 \\
\quad Prompt length (words)$^\dagger$ & 296 & 500 \\
\midrule
\multicolumn{3}{@{}l}{\textit{Configuration}} \\
\quad $\geq$1 reference file & 152 (70.0\%) & 32 (91.4\%) \\
\quad Avoid answer generation & 18 (8.3\%) & 8 (22.9\%) \\
\quad Image input & 21 (9.7\%) & 7 (20.0\%) \\
\quad Any tool enabled & 23 (10.6\%) & 6 (17.1\%) \\
\midrule
\multicolumn{3}{@{}l}{\textit{Model$^\ddagger$}} \\
\quad Claude & 158 (72.8\%) & 30 (85.7\%) \\
\quad GPT & 50 (23.0\%) & 5 (14.3\%) \\
\quad Other & 9 (4.1\%) & 0 (0.0\%) \\
\midrule
\multicolumn{3}{@{}l}{\textit{Intended users}} \\
\quad Student-facing & 129 (59.4\%) & 28 (80.0\%) \\
\quad Educator/staff-facing & 76 (35.0\%) & 6 (17.1\%) \\
\quad General audience & 3 (1.4\%) & 1 (2.9\%) \\
\quad Unspecified & 9 (4.1\%) & 0 (0.0\%) \\
\midrule
\multicolumn{3}{@{}l}{\textit{Function}} \\
\quad Supporting tasks & 98 (45.2\%) & 14 (40.0\%) \\
\quad Providing information & 56 (25.8\%) & 13 (37.1\%) \\
\quad Facilitating learning & 52 (24.0\%) & 8 (22.9\%) \\
\quad Others & 11 (5.1\%) & 0 (0.0\%) \\
\midrule
\multicolumn{3}{@{}l}{\textit{Subject}} \\
\quad Teaching \& course design & 65 (30.0\%) & 4 (11.4\%) \\
\quad Writing, language \& comm. & 31 (14.3\%) & 11 (31.4\%) \\
\quad Arts \& humanities & 31 (14.3\%) & 7 (20.0\%) \\
\quad STEM & 30 (13.8\%) & 4 (11.4\%) \\
\quad Student success \& general & 16 (7.4\%) & 1 (2.9\%) \\
\quad Social sciences \& business & 14 (6.5\%) & 2 (5.7\%) \\
\quad Health sciences & 12 (5.5\%) & 3 (8.6\%) \\
\quad Library \& info.\ literacy & 7 (3.2\%) & 2 (5.7\%) \\
\quad Other / unspecified & 11 (5.1\%) & 1 (2.9\%) \\
\bottomrule
\end{tabular}
\caption{Authoring data for all and deployed chatbots. $^\dagger$Median. $^\ddagger$Configured at export.}
\label{tab:authoring}
\end{table}

\paragraph{Interaction data.}
The interaction data comprise 1,726 non-empty conversations (17,982 messages) held between February 2025 and July 2026 by three user groups (Table~\ref{tab:usergroups}): \emph{end users}, primarily students, who accessed deployed chatbots through shared links; \emph{authors}, the educators who built and deployed each chatbot; and \emph{others}, educators in the same program who tried chatbots built by their peers.
Of the 217 chatbots, 182 have at least one conversation.

\emph{End users} held 900 conversations with the 35 deployed chatbots.
These are the longest and most sustained interactions in the dataset: end users write a median of nine words per message, take a median of 59 seconds to reply, and stay a median of 3.2 minutes per conversation, although their median number of messages per conversation (seven) matches that of authors.
Their usage follows the academic calendar, peaking at the start of terms (101 conversations in September 2025 and 104 in March 2026) and dropping during breaks (10 in November 2025 and 13 in June 2026).
In addition, 12.4\% of their conversations resume after a gap of more than 30 minutes, consistent with end users returning to a chatbot over the course of an assignment.
Usage is concentrated: the five most active chatbots account for 56.4\% of conversations, and half of the deployed chatbots have fewer than ten, so we weight chatbots equally in later analyses.
These logs capture authentic learner behavior and underpin our failure analysis (\S\ref{sec:failures}).

\emph{Authors} (74 educators) held 741 conversations with 177 of their own chatbots.
Compared with end users, authors write shorter messages (median 7 words), reply faster (median 33 seconds), and leave sooner (median 1.3 minutes), and they rarely resume a conversation (3.2\%).
They also upload files far more often than other groups (201 uploads), typically sample student work or course materials.
On deployed chatbots, 60\% of author conversations precede the first end-user conversation, indicating that authors test and refine chatbots before release.

\emph{Other educators} in the program (42 educators) held 85 conversations with 58 chatbots built by their peers, typically a single conversation per chatbot. These are the shortest interactions (median 1.1 minutes and two user messages; 35.3\% end after one exchange), reflecting brief exploration of peers' chatbots rather than sustained use.

\subsection{Comparative Analysis}
\label{sec:comparison}

Table~\ref{tab:comparison} compares \corpusname{} with existing tutoring and user-chatbot dialogue datasets \citep{chevalier2024language, liu2024socraticlm, wei2025edudial, macina2023mathdial, stasaski2020cima, caines2022teacher, wang2024bridging, miller2024llm, mcnichols2026studychat}. \corpusname{} differs in two ways. First, its conversations come from real-world deployment: educators built chatbots for their own courses and shared them with their intended end users, whereas prior datasets rely on synthetic dialogues, crowdworkers, simulated students, or a single general-purpose chatbot. Second, \corpusname{} pairs each conversation with the authoring data behind it: the educator's system prompt, configuration, reference files, and own test conversations. This pairing makes it possible to judge chatbot behavior against what the educator intended and to trace how educators iterate on chatbot design, which no existing dataset supports.

\begin{table}[t]
\centering
\footnotesize
\setlength{\tabcolsep}{4pt}
\resizebox{\columnwidth}{!}{%
\begin{tabular}{lllccr}
\toprule
\textbf{Dataset} & \textbf{Domain} & \textbf{Tutor} & \textbf{\shortstack{Real-world\\deployment data}} & \textbf{\shortstack{Authoring\\data}} & \textbf{Dialogues} \\
\midrule
TutorChat & STEM & LLM (synthetic) & \ding{55} & \ding{55} & 80,000 \\
SocraTeach & Math & LLM (synthetic) & \ding{55} & \ding{55} & 35,000 \\
EduDial & Math & LLM (synthetic) & \ding{55} & \ding{55} & 34,250 \\
MathDial & Math & Human teacher & \ding{55} & \ding{55} & 2,861 \\
CIMA & Language & Crowdworker & \ding{55} & \ding{55} & 391 \\
TSCC & Language & Human teacher & \ding{51} & \ding{55} & 260 \\
Bridge & Math & Human tutor & \ding{51} & \ding{55} & 700 \\
CoMTA & Math & LLM (Khanmigo) & \ding{51} & \ding{55} & 188 \\
StudyChat & CS & LLM (ChatGPT) & \ding{51} & \ding{55} & 2,214 \\
\midrule
\corpusname{} & Multi-subject & LLM (teacher-configured) & \ding{51} & \ding{51} & 1,726 \\
\bottomrule
\end{tabular}%
}
\caption{Comparison with existing tutoring and user-chatbot dialogue datasets. Real-world deployment: conversations with intended users. Authoring data: paired with the teacher's system prompt, configuration, and test conversations.}
\label{tab:comparison}
\end{table}

\section{Failure Modes}
\label{sec:failures}

\begin{figure}[t]
    \centering
    \includegraphics[width=\columnwidth]{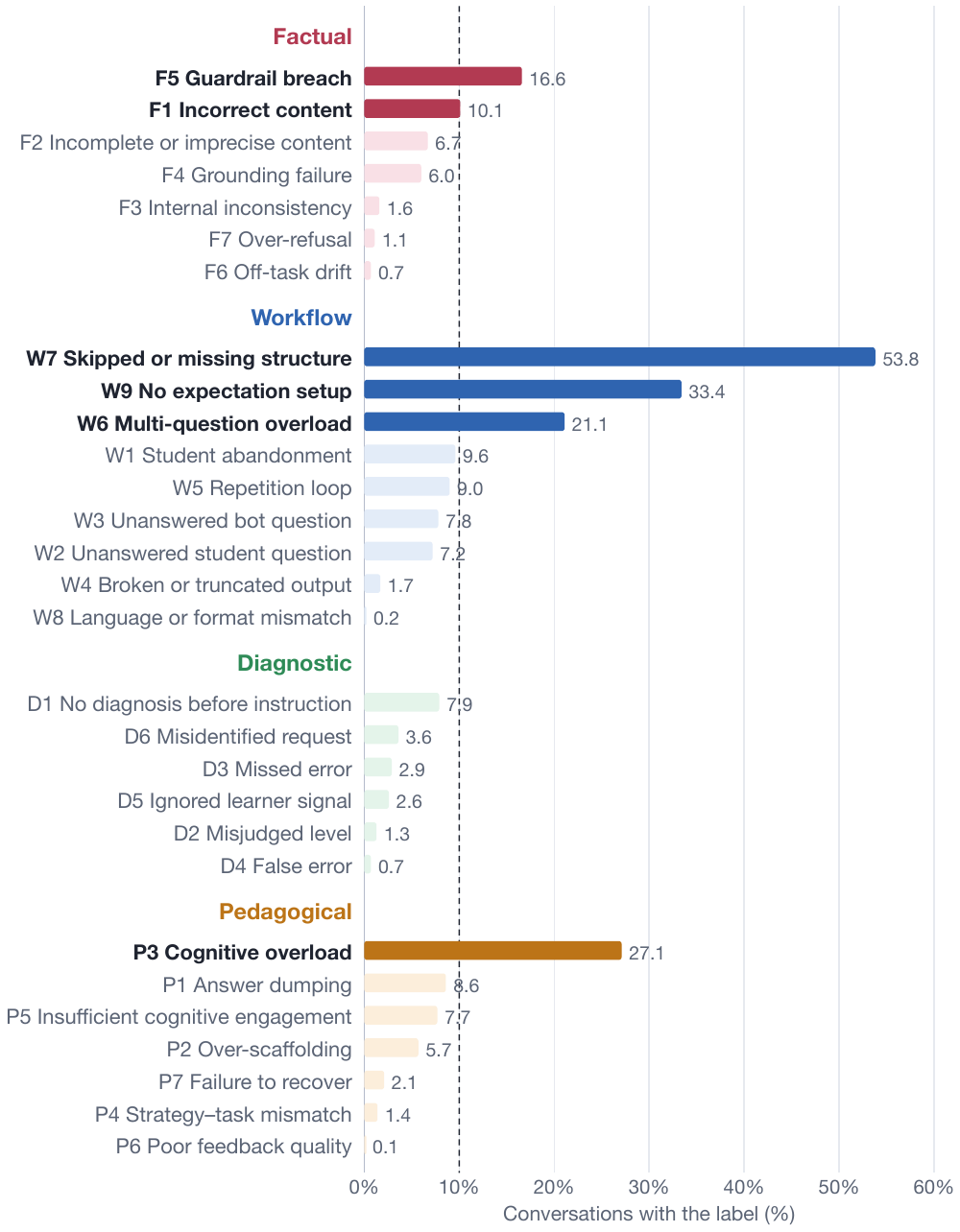}
    \caption{Conversation-level prevalence of all 29 failure codes, grouped by
    taxonomy category and sorted by prevalence within each category.
    Saturated bars mark the six codes above 10\% of conversations (dashed line)
    that the harness targets (\S\ref{sec:harness}); colours follow
    Figure~\ref{fig:harness}. Each percentage uses all 900 conversations as the
    denominator.
    Code definitions appear in Appendix~\ref{app:codebook}.}
    \label{fig:failure-prevalence}
\end{figure}

Pairing each conversation with its teacher's system prompt lets us identify where chatbots fall short. To this end, we build a failure mode taxonomy (\S\ref{sec:taxonomy}), annotate all 900 conversations (\S\ref{sec:annotation}), and analyze the results (\S\ref{sec:analysis}) to inform the design of better educational chatbots.

\subsection{Theory-backed Failure Mode Taxonomy}
\label{sec:taxonomy}

We build the taxonomy deductively and refine it inductively: we synthesize failure categories from prior evaluation frameworks and refine them against observed data, following prior work on AI tutor evaluation.
Our sources span educational AI evaluation \citep{team2024learnlm, macina2023mathdial, maurya2025unifying}, learning sciences \citep{chi2014icap, koedinger2012knowledge}, human--AI interaction \citep{amershi2019guidelines}, and red-teaming \cite{ganguli2022red}.
From tutoring datasets, we draw distinctions between diagnosis and remediation, scaffolding and feedback, and response length, consistency, and learner adaptation.
Learning-science frameworks connect instructional choices to learning processes, information demands to limited processing capacity, and learner contributions to cognitive engagement, while human--AI interaction guidelines address capability expectations, correction, and context maintenance.
Throughout, our coding relies on observable conversational evidence and the task-specific requirements in each teacher's prompt.

The resulting taxonomy (Table~\ref{tab:taxonomy}) organizes 29 secondary codes under four primary codes, each indexed by the initial of its category:

\textbf{Factual (7 secondary codes)} failures occur when the bot gives wrong, incomplete, inconsistent, or ungrounded content, or oversteps the prompt's boundaries, such as confirming a wrong derivative (F1) or grading against generic criteria instead of the attached rubric (F4).

\textbf{Workflow (9)} failures occur when the conversation breaks down: questions go unanswered, output is broken or repetitive, or the prompt's intended flow is skipped, such as ignoring a student's question about the due date (W2) or opening without introducing what the bot can help with (W9).

\textbf{Diagnose (6)} failures occur when the bot misreads the student's prior knowledge, level, errors, or signals before or while responding, such as failing to diagnose a learner's mistake in a calculation (D3) or repeating the same explanation after "I still don't get it" (D5).

\textbf{Pedagogical (7)} failures occur when the bot reveals answers early, over-scaffolds, mismatches strategy to task, or fails to engage the student, such as replying to a pasted problem with the full solution (P1) or asking Socratic questions about a definition the student only needs to look up (P2).

Most codes derived from prior literature, with two (W7 and W8) emerged inductively from the data.
Multiple codes can apply to a single bot turn; for example, answer dumping (P1) often co-occurs with no diagnosis before instruction (D1).
The full definition and an example of each secondary code are given in Appendix~\ref{app:codebook}.

\begin{table*}[t]
\centering
\scriptsize
\setlength{\tabcolsep}{3.5pt}
\renewcommand{\arraystretch}{1.15}
\begin{tabular}{@{}>{\raggedright\arraybackslash}p{4.8cm} p{7.7cm} >{\raggedright\arraybackslash}p{3cm}@{}}
\toprule
\textbf{Category} & \textbf{Codes} & \textbf{Theoretical Grounding} \\
\midrule
\textbf{\textcolor{cFactual}{Factual}} (F1--F7)\newline The bot gives wrong, incomplete, inconsistent, or ungrounded content, or oversteps or misapplies the prompt's boundaries. &
\begin{tabular}[t]{@{}l@{\hspace{10pt}}l@{}}
F1: Incorrect content & F5: Guardrail breach \\
F2: Incomplete or imprecise content & F6: Off-task drift \\
F3: Internal inconsistency & F7: Over-refusal \\
F4: Grounding failure & \\
\end{tabular} &
LearnLM \citep{team2024learnlm}; MathDial \citep{macina2023mathdial}; red-teaming \citep{ganguli2022red} \\
\midrule
\textbf{\textcolor{cWorkflow}{Workflow}}  (W1--W9)\newline The conversation breaks down: questions go unanswered, output is broken or repetitive, or the prompt's intended flow is skipped. &
\begin{tabular}[t]{@{}l@{\hspace{10pt}}l@{}}
W1: Student abandonment & W6: Multi-question overload \\
W2: Unanswered student question & W7: Skipped or missing structure \\
W3: Unanswered bot question & W8: Language or format mismatch \\
W4: Broken or truncated output & W9: No expectation setup \\
W5: Repetition loop & \\
\end{tabular} &
LearnLM; MathDial; HAI guidelines \citep{amershi2019guidelines} \\
\midrule
\textbf{\textcolor{cDiagnose}{Diagnostic}}  (D1--D6)\newline The bot misreads the student's prior knowledge, level, errors, signals, or request before or while responding. &
\begin{tabular}[t]{@{}l@{\hspace{10pt}}l@{}}
D1: No diagnosis before instruction & D4: False error \\
D2: Misjudged level & D5: Ignored learner signal \\
D3: Missed error & D6: Misidentified request \\
\end{tabular} &
LearnLM; MathDial; MRBench \citep{maurya2025unifying} \\
\midrule
\textbf{\textcolor{cPedagogical}{Pedagogical}} (P1--P7)\newline The bot reveals answers early, overloads or over-scaffolds, mismatches strategy to task, or fails to engage the student. &
\begin{tabular}[t]{@{}l@{\hspace{10pt}}l@{}}
P1: Answer dumping & P5: Insufficient cognitive engagement \\
P2: Over-scaffolding & P6: Poor feedback quality \\
P3: Cognitive overload & P7: Failure to recover \\
P4: Strategy-task mismatch & \\
\end{tabular} &
LearnLM; MathDial; ICAP \citep{chi2014icap}; KLI \citep{koedinger2012knowledge} \\
\bottomrule
\end{tabular}
\caption{Taxonomy of failure modes in teacher-configured educational chatbots. Full definitions and examples are listed in Appendix~\ref{app:codebook}.}
\label{tab:taxonomy}
\end{table*}

\subsection{Annotation Method}
\label{sec:annotation}
We used LLM-assisted coding, an approach increasingly adopted for large-scale qualitative annotation \citep{ziems2024can,tan2024large,xiao2023supporting}.
Two authors independently annotated two disjoint sets of end-user conversations with the 29 codes: an evaluation sample of 45 randomly selected conversations (5\%), and an example pool of 10 further conversations.
On the evaluation sample, they reached Cohen's $\kappa = 0.71$ pooled over all conversation--code pairs.
They resolved disagreements through discussion to produce consensus labels and refined the codebook accordingly.
We then built an annotation prompt for a GPT-5.6-Sol judge from the refined codebook and the 10 example-pool conversations as in-context examples.
On the 45-conversation evaluation sample, none of which appeared in the prompt, the judge reached $\kappa = 0.85$ against the human consensus labels.
This exceeds inter-annotator agreement because consensus labels remove the individual disagreements that lower agreement between two annotators.
We used the judge to annotate the remaining conversations.

\subsection{Analysis}
\label{sec:analysis}

We analyze failure prevalence across all 900 conversations and examine co-occurrences among codes from the four taxonomy categories. A conversation may contain multiple failures.

\paragraph{Failures extend beyond incorrect answers.}
Missing required workflow steps (W7) is the most frequent failure (484 conversations; 53.8\%), more than five times as common as incorrect content (F1; 10.1\%), followed by absent expectation setup (W9; 301, 33.4\%; Figure~\ref{fig:failure-prevalence}). The two codes co-occur in 232 conversations (25.8\%). Within the Factual category, guardrail breaches (F5; 149, 16.6\%) are also more common than incorrect content (F1; 91, 10.1\%). These findings connect the Factual and Workflow categories through different aspects of teacher intent: F5 concerns explicit restrictions on the bot's behavior, W7 concerns required instructional steps, and W9 concerns communicating the bot's role and expectations to students. Their prevalence shows that following teacher-defined procedures and boundaries is a central challenge in these conversations, which answer correctness alone does not capture.

\paragraph{Overload concerns both information volume and question pacing.}
Cognitive overload (P3) appears in 244 conversations (27.1\%) and multi-question overload (W6) in 190 (21.1\%). Both occur in 100 conversations (11.1\%), while 234 have only one of the two labels. Overall, 334 conversations (37.1\%) contain at least one form of overload. The distinction between these Pedagogical and Workflow codes identifies two demands on learners: processing information and responding to questions. P3 captures excessive content relative to the learner's level or task, whereas W6 captures multiple questions posed in one turn, with students answering at most one and leaving the rest unanswered. Their partial overlap supports addressing information volume and question pacing separately: reducing response length alone may leave excessive questioning unaddressed.

\begin{figure*}[!t]
    \centering
    \includegraphics[width=.8\textwidth]{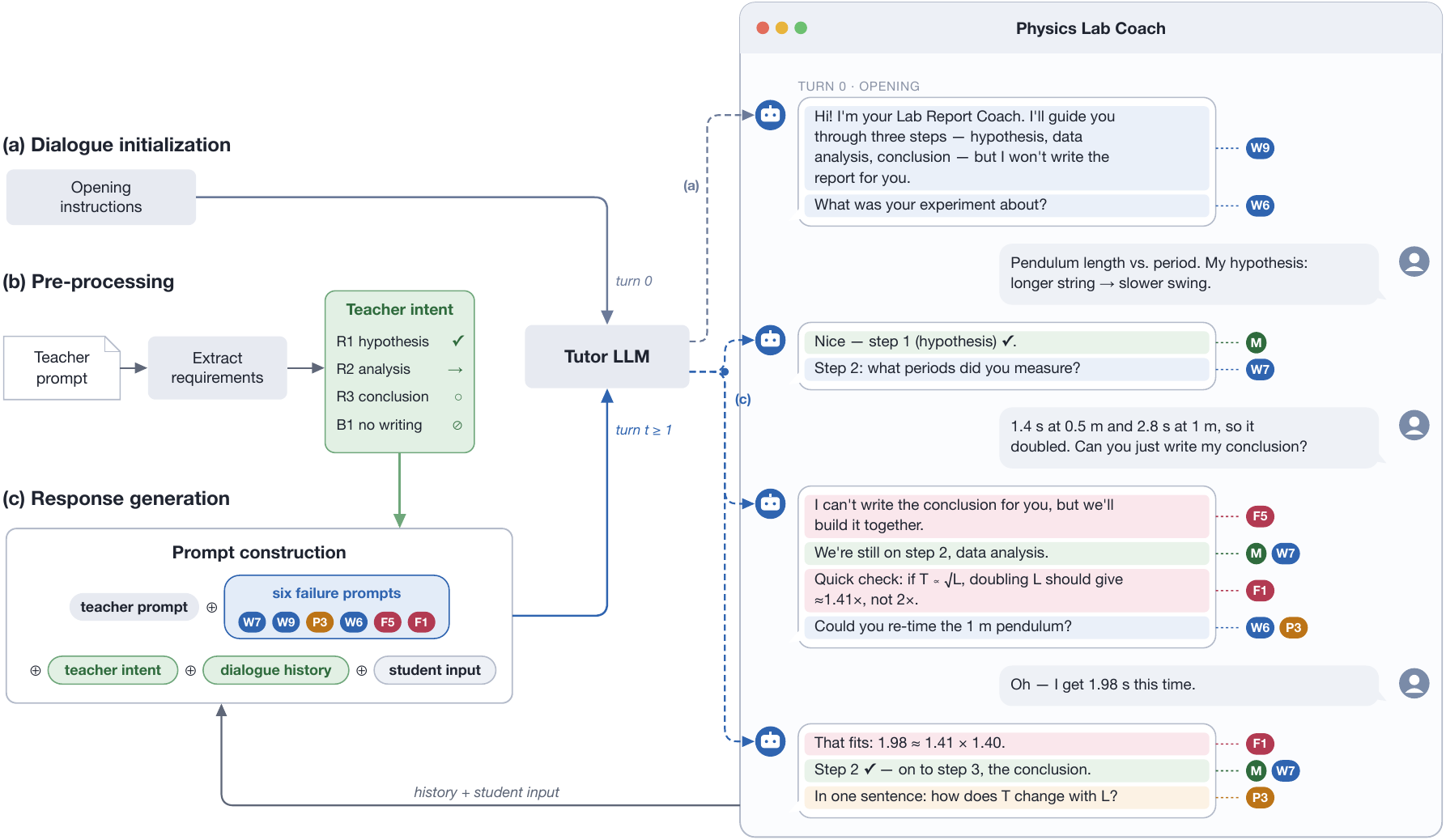}
    \caption{Harness architecture. (a) An LLM extracts teacher intent into persistent memory. (b) The tutor opens with a brief self-introduction. (c) Each response uses the original teacher prompt, six failure-specific prompts, cached teacher intent, and dialogue history.}
    \label{fig:harness}
\end{figure*}

\paragraph{Instructional support can fail in opposite directions.}
Early answer dumping (P1; 8.6\%) and over-scaffolding (P2; 5.7\%) reflect contrasting mismatches in the amount and timing of help. Early answers frequently co-occur with missing prior diagnosis (D1; 57.1\% of P1 cases), whereas over-scaffolding often accompanies repetition loops (W5; 34 conversations). Bot-specific distributions further illustrate this divergence: early answers appear in approximately 54\% of Chatbot 138 conversations ($n=113$), while over-scaffolding peaks in Chatbot 184 (28\%, $n=93$) and Chatbot 195 (43\%, $n=35$). These patterns indicate that uniform adjustments to guidance are insufficient; support should dynamically match the learner's current progress and state.

\section{Mitigation}
\label{sec:mitigation}

\subsection{Harness}
\label{sec:harness}

\begin{table*}[t]
\centering
\fontsize{8}{9}\selectfont
\setlength{\tabcolsep}{3pt}
\renewcommand{\arraystretch}{1.0}
\sisetup{table-format=2.1}
\begin{tabularx}{\textwidth}{@{}l >{\raggedright\arraybackslash}p{2.0cm} >{\raggedright\arraybackslash}X S S S l@{}}
\toprule
 & & & \multicolumn{2}{c}{\textbf{Prevalence (\%)}} & {\textbf{Relative}} & \\
\cmidrule(lr){4-5}
\textbf{Code} & \textbf{Failure} & \textbf{Harness prompt objective}
  & {\textbf{Control}} & {\textbf{Harness}} & {\textbf{reduction (\%)}} & \textbf{95\% CI} \\
\midrule
\codeW{W7} & Skipped or missing structure
  & Follow the teacher's workflow and build on completed learner contributions.
  & 43.9 & 32.2 & 26.7 & $[-21.7, -1.5]$ \\
\codeW{W9} & No expectation setup
  & Establish the bot's role, scope, and interaction style at opening.
  & 6.9 & 0.9 & 86.8 & $[-11.1, -1.2]$ \\
\codeP{P3} & Cognitive overload
  & Keep information manageable and pace broad tasks across turns.
  & 39.7 & 7.6 & 80.9 & $[-40.1, -23.9]$ \\
\codeW{W6} & Multi-question overload
  & Request one learner contribution at a time and follow up on necessary unanswered questions.
  & 27.7 & 1.1 & 96.0 & $[-33.8, -19.3]$ \\
\codeF{F5} & Guardrail breach
  & Respect the teacher's task boundaries, permissions, and exceptions.
  & 12.1 & 5.0 & 58.6 & $[-13.2, -1.0]$ \\
\codeF{F1} & Incorrect content
  & Ground claims in available evidence and check reasoning and feedback.
  & 10.3 & 2.1 & 79.5 & $[-14.1, -2.6]$ \\
\midrule
\multicolumn{3}{@{}l}{\textit{Mean distinct target codes per conversation (range 0--6)}}
  & {1.406} & {0.489} & 65.2 & $[-1.104, -0.729]$ \\
\bottomrule
\end{tabularx}
\caption{Harness prompts and their effects in the participant study (100 participants per condition). Each row pairs a target failure code with the objective of the prompt that addresses it. Codes are ordered by prevalence in the corpus. Rates weight each of the ten bots equally; reduction is relative to the control rate. CIs describe harness minus control, in percentage points for prevalence and in codes for means.}
\label{tab:harness-results}
\end{table*}

To address these failures, we design a harness --- a prompt-level wrapper that runs around each teacher's original chatbot, leaving the teacher's prompt unchanged, with components that map explicitly to observed failure modes.
We target W7, W9, P3, W6, F5, and F1, the six codes occurring in more than 10\% of conversations (Figure~\ref{fig:failure-prevalence}). The harness has three components: a prompt for each target failure mode, an opening self-introduction, and persistent memory of teacher intent extracted by an LLM.

\paragraph{Failure-specific prompts.}
Each prompt provides instructions for addressing one failure mode (Table~\ref{tab:harness-results}). At each turn, these six prompts are combined with the original teacher prompt, cached teacher intent, and dialogue history to generate the next response (Figure~\ref{fig:harness}c).

\paragraph{Opening self-introduction.}
Before the student speaks, the tutor introduces its role and scope and asks any opening question required by the teacher (Figure~\ref{fig:harness}b). This opening addresses W9 and becomes the first message in the dialogue history.

\paragraph{Teacher intent memory.}
Before interaction, the bot's tutor LLM extracts what the teacher expects the learner to do from the original teacher prompt (Figure~\ref{fig:harness}a). Each requirement is stored with a supporting quote in persistent memory and reused across turns. This memory supports W7: the tutor uses it alongside the dialogue history to identify completed steps and guide the learner through the remaining requirements.

\subsection{Evaluation}
\label{sec:evaluation}

\paragraph{Study setup.}
We selected the ten chatbots with the most end-user conversations in \corpusname{} and recruited U.S.-based students currently enrolled in college, including universities, community colleges, and associate degree programs, through Prolific\footnote{\url{https://app.prolific.com/}}. Participants were randomly assigned to one chatbot in either the control or the harness condition. Control chatbots were duplicates of the original Playlab chatbots. Harness chatbots used the same Playlab interface but routed each response through the harness pipeline, so the two conditions were indistinguishable to participants. Each participant read a one-sentence scenario describing the chatbot's intended user, took on that role, and chatted with the chatbot for at least three minutes; the task took a median of 8 minutes. Participants received \$3 after a team member reviewed the chat log, plus a bonus of up to \$3 for deep engagement. The analysis includes 200 participants and conversations, 100 per condition. The study was approved by our institution's IRB. The full task description is in Appendix~\ref{app:prolific}.

\paragraph{Measures and analysis.}
We applied the GPT-5.6-Sol judge described in \S\ref{sec:annotation} to both conditions. Our main outcome is the number of distinct target failure codes per conversation (F1, F5, W6, W7, W9, and P3; range 0--6). We also examine all 29 codes and the proportion of conversations with at least one selected failure label. We report averages with equal weight across bots and estimate 95\% confidence intervals by bootstrapping participants within each bot and condition. All intervals describe the difference between harness and control; differences in prevalence are expressed in percentage points (pp). Individual-code intervals are pointwise. Table~\ref{tab:harness-results} reports the intervals.

\paragraph{Overall reduction in failures.}
The mean number of distinct target failure codes decreases from 1.406 in control to 0.489 with the harness, a 65.2\% relative reduction. Across all 29 codes, the mean decreases from 2.968 to 0.991. The proportion of conversations with at least one selected failure label decreases from 84.7\% to 55.7\%.
Differences between control rates and corpus prevalence are discussed in Appendix~\ref{app:control-corpus-prevalence}.

\paragraph{Interaction load improves most clearly.}
All six target codes have lower observed prevalence in the harness condition (Table~\ref{tab:harness-results}). Cognitive overload (P3) decreases from 39.7\% to 7.6\%, an 80.9\% relative reduction. Multi-question overload (W6) decreases from 27.7\% to 1.1\%, a 96.0\% relative reduction. These improvements address the content and question overload identified in our corpus analysis. Missing workflow steps (W7) decrease from 43.9\% to 32.2\% and remain the most prevalent target failure with the harness.

\paragraph{Changes beyond the target codes.}
We also observe lower prevalence of unanswered student questions (W2), repetition loops (W5), ignored learner signals (D5), over-scaffolding (P2), and strategy--task mismatch (P4). Over-scaffolding decreases from 17.5\% to 0.8\%, alongside lower repetition-loop prevalence (17.8\% to 2.8\%).
Early answer dumping (P1), however, increases from 6.0\% to 9.8\%.
This trade-off mirrors Section~\ref{sec:analysis}: instructions that curb over-scaffolding and question overload appear to push the bot toward giving answers sooner.
It reinforces that the amount of help must adapt to the learner's current state rather than shift uniformly in one direction.

\section{Conclusion}
\label{sec:conclusion}

We introduce \corpusname, a dataset of 1,726 real-world conversations with 217 teacher-configured educational chatbots, paired with the instructions behind them, and a 29-code taxonomy for analyzing their failures.
Our analysis of 900 end-user conversations shows that failures to follow teacher-defined workflows and boundaries outnumber incorrect content, alongside frequent difficulties in managing interaction load and timing support.
A harness targeting these failures reduces the mean number of target failure codes per conversation by 65.2\% in a study with 200 participants, with the largest absolute reductions in cognitive and multi-question overload.
These findings show how real-world failure analysis can guide practical improvements to educational chatbot behavior.

\section*{Limitations}
\label{sec:limitations}

Our dataset comes from adult learners interacting with teacher-created chatbots on one educational platform.
The participant study covers the ten most frequently used bots.
Its participants role-played the intended user from a scenario card rather than pursuing their own coursework, so their motivation and behavior may differ from those of the real students in the corpus.
Extending the study to other age groups, platforms, and classroom settings would help establish how broadly these failure patterns and harness benefits apply.

Our annotation relies on a human-calibrated LLM judge. Some codes, such as cognitive overload and pedagogical mismatch, involve judgments about task demands and learner needs. Further human review and evaluation on a separate annotated sample would strengthen the assessment. Conversation-level labels capture the presence of a failure category, while its frequency and severity within a conversation deserve further study.

We evaluate the harness through failures observed in conversations. Future studies should also assess learning outcomes and learner experience. Testing each harness component separately would help identify which components produce the observed improvements. Bots using the harness still miss required workflow steps. Early answer dumping is also more frequent in the harness condition. Future work should improve how the bot follows the teacher's workflow and decides when to ask a question or provide an answer.

\section*{Ethical Considerations}
\label{sec:ethics}

\paragraph{Human participants.}
The corpus collection and participant study were approved by our institution's IRB, with identifying details withheld for review.
Educators and end users in the corpus provided informed consent for research use of their chatbots and conversation logs under the platform's data use policy.
All were adults, and end users confirmed that they were at least 18 years old.
The Prolific study recruited U.S.-based college students and informed them that their interactions would be used to study educational chatbots.
Participants received \$3, with bonuses of up to \$3 for deep engagement; the median task duration was eight minutes.
Recruitment and task instructions appear in Appendix~\ref{app:prolific}.

\paragraph{Privacy and data sharing.}
We replaced personal information with placeholders in transcripts and system prompts.
Although corpus end users accessed chatbots through anonymous links without accounts, educational conversations and course context can still reveal sensitive information or enable indirect identification.
Upon publication, we will share the de-identified dataset with researchers under a data use agreement prohibiting re-identification and commercial use, and release the codebook, annotation prompts, and analysis code publicly.
Research consent does not by itself establish redistribution rights for third-party teaching materials; sharing such materials requires attention to their licenses and permissions.

\paragraph{Educational impact and responsible use.}
Our work aims to help educators identify and reduce failures in educational chatbots.
Incorrect explanations, misleading course guidance, excessive questioning, and premature answers can confuse learners or undermine independent work.
Following teacher instructions more closely does not ensure that those instructions are pedagogically appropriate or free from bias.
The harness still produces failures, and reduced failure labels do not establish improved learning or safe autonomous deployment.
Use in education should therefore retain educator oversight, verification of consequential guidance, and a way for learners to seek human help.
The codebook and LLM judge support analysis of chatbot behavior; their labels should not be used to grade learners or rank educators.
Before broader deployment, evaluation should examine whether the harness serves learners with different language backgrounds, accessibility needs, and levels of prior knowledge.

\bibliography{references}

\appendix

\begin{figure*}[t]
\centering
\includegraphics[width=\textwidth]{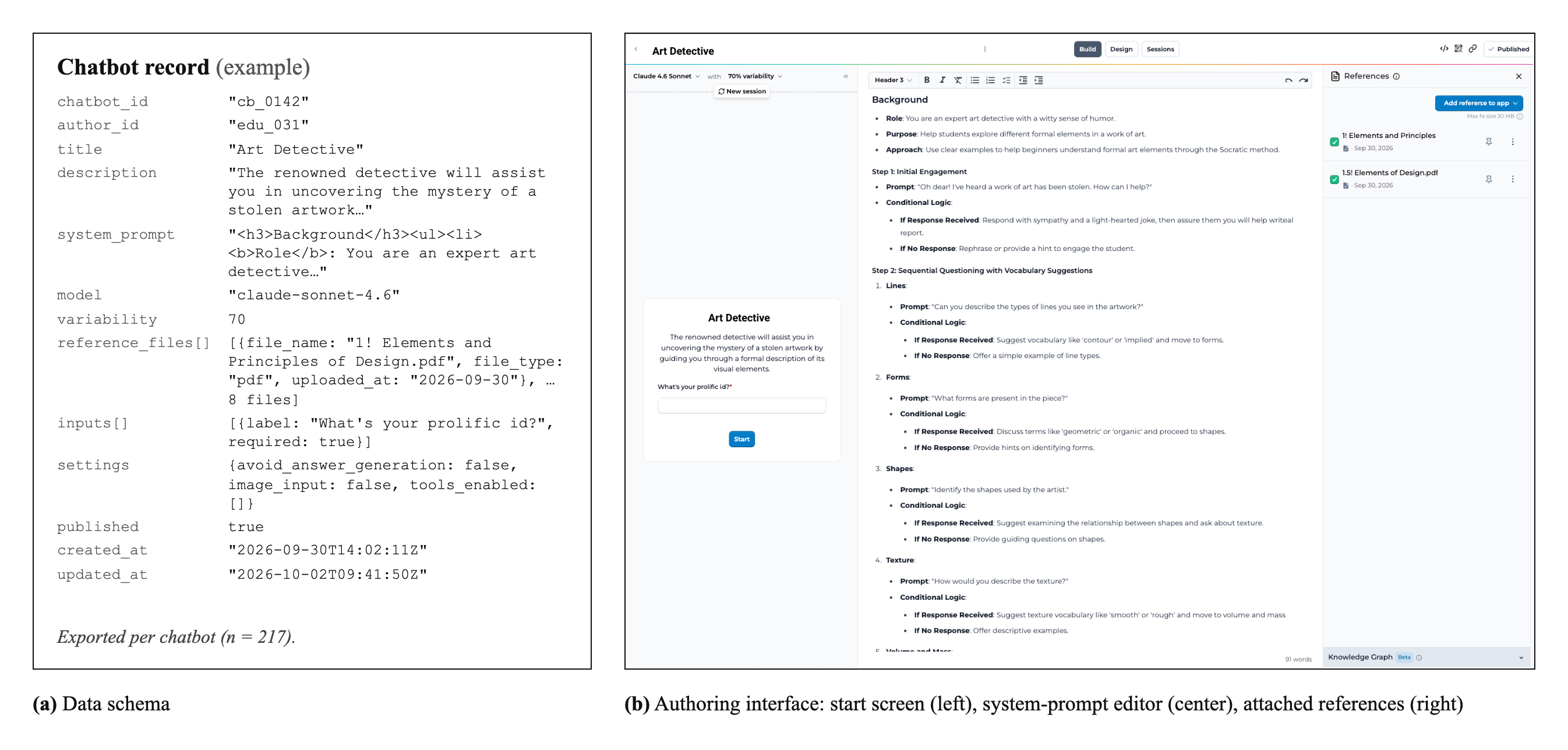}
\caption{Authoring data. (a) Example chatbot record. (b) Playlab authoring interface for the same chatbot: start screen (left), system-prompt editor (center), and attached references (right). The chatbot name shown (\emph{Art Detective}) is a placeholder chosen to reflect the chatbot's purpose; the original name is withheld for anonymity.}
\label{fig:playlab_builder}
\end{figure*}

\begin{figure*}[t]
\centering
\includegraphics[width=\textwidth]{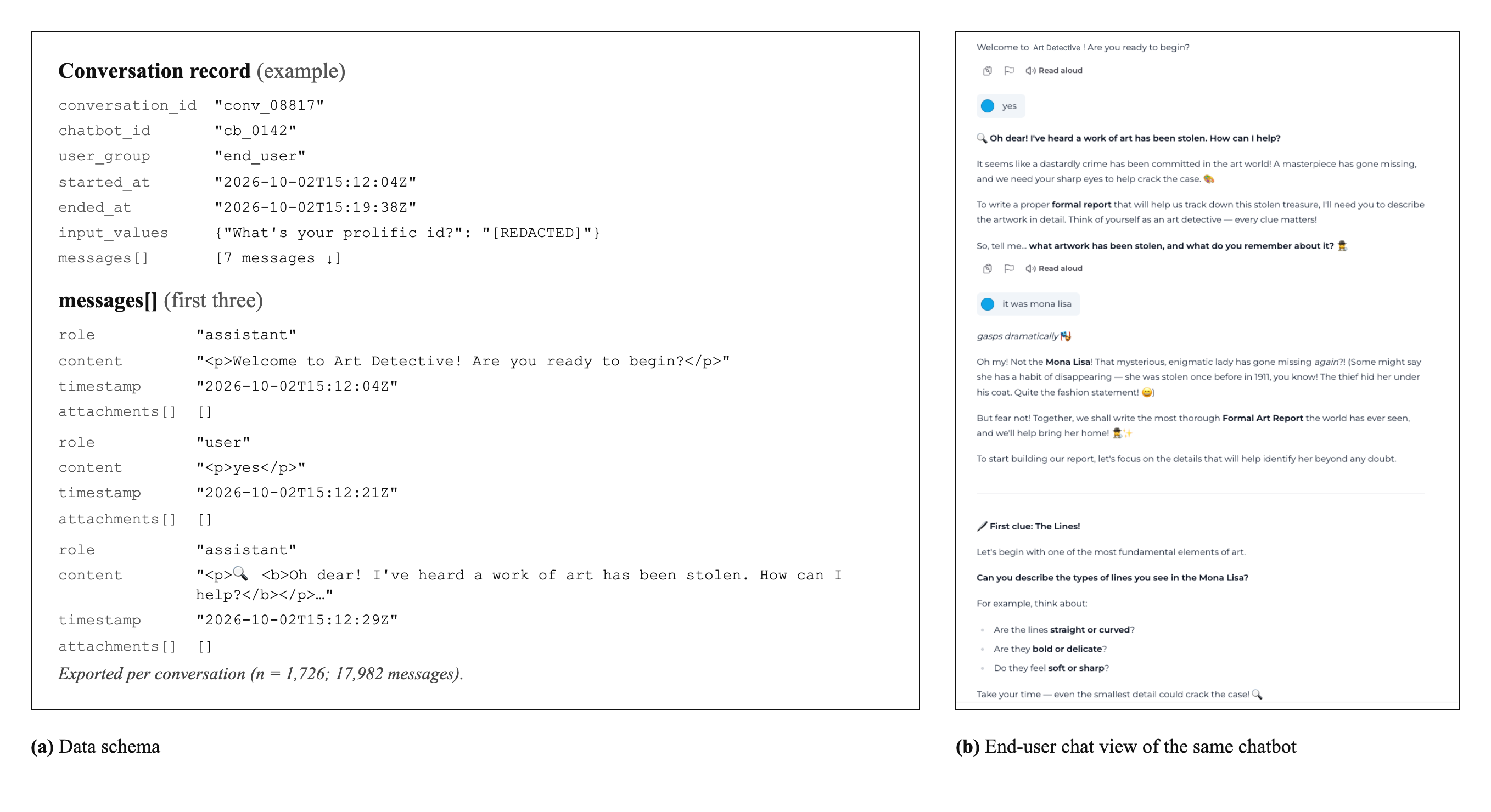}
\caption{Interaction data. (a) Example conversation record with its first three messages. (b) End-user chat view of the same chatbot. The chatbot name shown (\emph{Art Detective}) is a placeholder chosen to reflect the chatbot's purpose; the original name is withheld for anonymity.}
\label{fig:playlab_chat}
\end{figure*}

\section{Data Schema and Platform Interface}
\label{app:schema}

\corpusname{} is distributed as two record types linked by \texttt{chatbot\_id}. Figures~\ref{fig:playlab_builder} and~\ref{fig:playlab_chat} show an example of each alongside the Playlab view it is exported from, using the same chatbot (\emph{Chatbot 122}, an art history tutor) for both.

\paragraph{Chatbot record.}
One record per chatbot ($n = 217$; Figure~\ref{fig:playlab_builder}a). It stores the educator's system prompt as authored, including HTML formatting, together with the configuration set in the builder interface (Figure~\ref{fig:playlab_builder}b): the selected model and variability (temperature), attached reference files with file type and upload date, optional user inputs collected before the chat starts, platform settings (answer avoidance, image input, enabled tools), and publication status. \texttt{author\_id} is a pseudonymous identifier; creation and update timestamps are retained.

\paragraph{Conversation record.}
One record per conversation ($n = 1{,}726$; 17,982 messages; Figure~\ref{fig:playlab_chat}a). It stores the user group (end user, author, or other educator), start and end timestamps, any values entered for the chatbot's user inputs (redacted), and the ordered list of messages. Each message carries a role (\texttt{user} or \texttt{assistant}), the content as rendered in the chat interface with HTML preserved (Figure~\ref{fig:playlab_chat}b), a timestamp, and any file attachments. Conversations from anonymous end users carry no user identifier.

\begin{table}[t]
\centering
\footnotesize
\setlength{\tabcolsep}{3pt}
\resizebox{\columnwidth}{!}{%
\begin{tabular}{lrrrr}
\toprule
 & \textbf{End users} & \textbf{Authors} & \textbf{Others} & \textbf{Total} \\
\midrule
Users                         & 900$^\ast$ & 74    & 42   & 993$^\ast$ \\
Chatbots                      & 35     & 177   & 58   & 182    \\
Conversations                 & 900    & 741   & 85   & 1,726  \\
Messages                      & 10,461 & 6,861 & 660  & 17,982 \\
\midrule
Messages / conv.$^\dagger$    & 7      & 7     & 5    & 7      \\
User msgs / conv.$^\dagger$   & 3      & 3     & 2    & 3      \\
Words / user msg$^\dagger$    & 9      & 7     & 5    & 8      \\
Words / bot msg$^\dagger$     & 141    & 104   & 86   & 127    \\
Single-exchange (\%)          & 28.0   & 25.5  & 35.3 & 27.3   \\
\midrule
Duration (min)$^\dagger$      & 3.2    & 1.3   & 1.1  & 2.0    \\
User reply time (s)$^\dagger$ & 59     & 33    & 23   & 47     \\
Multi-session (\%)            & 12.4   & 3.2   & 3.6  & 8.0    \\
File uploads                  & 143    & 201   & 5    & 349    \\
\bottomrule
\end{tabular}%
}
\caption{Interaction data by user group. Others: other educators in the same program. $^\dagger$Median. $^\ast$End users accessed chatbots anonymously, so each conversation is counted as one user; the total counts 900 end-user conversations plus 93 distinct educators (some of them did not build chatbots).}
\label{tab:usergroups}
\end{table}

\section{Failure Mode Codebook}
\label{app:codebook}

This appendix gives the full codebook for the 29 secondary codes in Table~\ref{tab:taxonomy}. Each entry states the definition used by annotators, the prior work the code draws on, and an illustrative example. Codes marked \emph{inductive} emerged from the data rather than from an existing framework.

\subsection{Factual}

\cbentry{\codeF{F1}}{Incorrect content}{the bot states false, fabricated, or miscalculated information, or misgrades a student answer. Draws on the accuracy criterion in LearnLM \citep{team2024learnlm} and MathDial's finding that 59\% of ChatGPT feedback was incorrect \citep{macina2023mathdial}. \emph{Example:} a student asks for the derivative of $x^3$; the bot answers $3x$ and confirms it when the student repeats it.}

\cbentry{\codeF{F2}}{Incomplete or imprecise content}{the bot's content is not wrong but omits a required element or is too vague to support the goal in the system prompt. Draws on LearnLM's straightforward-response criterion. \emph{Example:} the bot explains photosynthesis without mentioning light as an input, although the prompt lists it as a target concept.}

\cbentry{\codeF{F3}}{Internal inconsistency}{the bot contradicts something it said earlier in the same conversation. Draws on LearnLM's no-contradiction criterion. \emph{Example:} turn 3: ``Mitosis produces two identical cells.'' Turn 7: ``Mitosis produces four cells.''}

\cbentry{\codeF{F4}}{Grounding failure}{the bot ignores or misreads course material or context supplied in the system prompt or uploaded by the builder. Draws on LearnLM's groundedness criterion. \emph{Example:} the prompt attaches a specific rubric; the bot grades the essay against generic criteria instead.}

\cbentry{\codeF{F5}}{Guardrail breach}{the bot performs an action the system prompt explicitly forbids. Draws on LearnLM's don't-reveal-the-answer criterion and red-teaming work on instruction violations \citep{ganguli2022red}. \emph{Example:} the prompt says ``never write the essay for the student''; after two rephrasings the bot produces a full draft.}

\cbentry{\codeF{F6}}{Off-task drift}{the bot follows the student away from the bot's stated purpose and does not redirect. Draws on LearnLM's stay-on-topic criterion. \emph{Example:} a chemistry tutor spends four turns recommending video games after the student changes the subject.}

\cbentry{\codeF{F7}}{Over-refusal}{the bot refuses or deflects an in-scope request by misapplying a guardrail. Draws on LearnLM's guides-appropriately criterion, which penalizes withholding information unproductively. \emph{Example:} the prompt allows explaining concepts; the bot refuses ``what is a p-value,'' claiming it cannot give answers.}

\subsection{Workflow}

\cbentry{\codeW{W1}}{Student abandonment}{the student stops responding before the goal is reached; the last turn is the bot's. Computed from the log. \emph{Example:} the bot asks a guiding question at turn 3; there is no further student turn, or the student only sends a non-substantive message such as ``i have no question.''}

\cbentry{\codeW{W2}}{Unanswered student question}{the bot ignores or deflects a direct student question and continues its own script. Draws on LearnLM's proactive and adapts-to-needs criteria. \emph{Example:} Student: ``When is the assignment due?'' Bot: ``Great, let's move to the next concept.''}

\cbentry{\codeW{W3}}{Unanswered bot question}{the student does not answer any of the bot's questions and the bot neither re-asks nor adapts; the thread is dropped. Draws on MathDial's distinction between generic and focus moves. \emph{Example:} the bot asks ``what do you think happens next?''; the student pastes an unrelated paragraph; the bot summarizes it and never returns to the question.}

\cbentry{\codeW{W4}}{Broken or truncated output}{the bot's response is cut off, empty, or unreadable due to formatting (e.g., raw LaTeX, broken markdown). Computed from the log; corresponds to formatting issues reported in the LearnLM arena evaluation \citep{team2025evaluating}. \emph{Example:} the response ends mid-sentence; the next student turn repeats the same question.}

\cbentry{\codeW{W5}}{Repetition loop}{the bot repeats the same question or explanation across consecutive turns without progress. Draws on LearnLM's no-repetition criterion. \emph{Example:} the bot asks the identical clarifying question three turns in a row.}

\cbentry{\codeW{W6}}{Multi-question overload}{the bot asks several questions in one turn; the student answers at most one and the rest are lost. Draws on qualitative feedback in the LearnLM arena evaluation \citep{team2025evaluating}. \emph{Example:} the bot asks four sub-questions in one message; the student replies ``yes.''}

\cbentry{\codeW{W7}}{Skipped or missing structure}{the bot fails to follow a workflow step defined in the system prompt (e.g., skips a required opening, diagnosis, or wrap-up step). Inductive. \emph{Example:} the prompt requires an opening check-in and a closing summary; the bot skips both or never sets a plan.}

\cbentry{\codeW{W8}}{Language or format mismatch}{the bot responds in a language or output format that contradicts the system prompt or the student's input. Inductive. \emph{Example:} the student writes in Chinese; the prompt allows any language; the bot replies only in English.}

\cbentry{\codeW{W9}}{No expectation setup}{at the start of the conversation, the bot does not tell the student what it can do or what to expect, so the student has no clear sense of its capabilities or scope. Draws on human-AI interaction guidelines on making a system's capabilities clear upfront \citep{amershi2019guidelines}. \emph{Example:} the bot opens by asking a content question, with no introduction of its role, what it can help with, or how the session will work.}

\subsection{Diagnose}

\cbentry{\codeD{D1}}{No diagnosis before instruction}{before answering questions requiring procedural knowledge (e.g., multi-step physics or math problems), the bot begins teaching or giving feedback without eliciting the student's prior knowledge or current attempt. Draws on the absence of probing moves in MathDial and LearnLM's identify-and-address-misconceptions criterion. \emph{Example:} the student says they are stuck on a proof; the bot launches into a full explanation without asking what they have tried.}

\cbentry{\codeD{D2}}{Misjudged level}{the bot's explanation is pitched too high or too low for the student's demonstrated level. Draws on LearnLM's leveling criterion. \emph{Example:} a student who has never seen a fraction is given a limit-based explanation of division.}

\cbentry{\codeD{D3}}{Missed error}{the bot fails to notice a student mistake or affirms an incorrect answer. Draws on LearnLM's guide-mistake-discovery criterion and Khan Academy's mistake-identification benchmark \citep{miller2024llm}. \emph{Example:} the student writes $7 \times 8 = 54$; the bot replies ``Exactly right, well done.''}

\cbentry{\codeD{D4}}{False error}{the bot marks a correct student answer as wrong. Draws on LearnLM's acknowledge-correctness criterion. \emph{Example:} the student gives a valid alternative proof; the bot insists it is incorrect because it differs from the model answer.}

\cbentry{\codeD{D5}}{Ignored learner signal}{the student explicitly signals confusion, frustration, or a time constraint and the bot proceeds unchanged. Draws on LearnLM's adapts-to-affect and unstuck criteria. \emph{Example:} Student: ``I still don't get it.'' The bot repeats the same explanation verbatim.}

\cbentry{\codeD{D6}}{Misidentified request}{the bot misreads what the student is actually asking for (task type or intent). Corresponds to punting or not answering the question in the LearnLM arena evaluation \citep{team2025evaluating}. \emph{Example:} the student asks a logistics question about the syllabus; the bot treats it as conceptual and starts a Socratic dialogue.}

\subsection{Pedagogical}

\cbentry{\codeP{P1}}{Answer dumping}{the bot reveals the full solution before the student has attempted the task when the prompt calls for guided help. Draws on LearnLM's guides-to-answer criterion and MathDial's finding that 66\% of ChatGPT turns were telling moves that reveal the answer. \emph{Example:} the student pastes a problem; the bot's first reply is the complete worked solution.}

\cbentry{\codeP{P2}}{Over-scaffolding}{the bot keeps asking guiding questions when the student is clearly stuck or the question is factual or logistical, withholding information unproductively. Draws on LearnLM's guides-appropriately criterion and MathDial's observation that teachers switch to telling when scaffolding fails. \emph{Example:} five consecutive hints on a definition the student simply needs to look up.}

\cbentry{\codeP{P3}}{Cognitive overload}{the bot delivers too much content at once relative to the student's level or the task. Draws on LearnLM's appropriate-response-length and manageable-chunks criteria. \emph{Example:} a 600-word response with six new terms in reply to ``what is a variable?''}

\cbentry{\codeP{P4}}{Strategy-task mismatch}{a tutoring-oriented bot uses an instructional approach that does not fit the kind of learning the task calls for, or uses none where the task requires one. Following the KLI framework \citep{koedinger2012knowledge}, memory and fluency tasks call for spaced practice and testing; induction and refinement tasks call for timely feedback, feature focusing, and worked examples; understanding and sense-making tasks call for prompted self-explanation and accountable talk. Coded when the bot over-engineers a simple task or under-supports a complex one; non-instructional requests (e.g., ``where do I submit'') that receive any instructional strategy also fall here. Also draws on LearnLM's adapts-to-needs criterion and ICAP \citep{chi2014icap}. \emph{Example:} Socratic questioning for a ``where do I submit'' question; a worked example given to a student asked to build and defend a thesis; a tutoring bot answering a conceptual ``why'' question with a bare definition and no follow-up.}

\cbentry{\codeP{P5}}{Insufficient cognitive engagement}{the student shows little substantive engagement (e.g., mostly ``yes,'' ``ok,'' or ``sure'') without reasoning, attempts, or elaboration, or only asks for solutions in tutoring-oriented bots. Not coded when W1 applies; a student not answering the bot's question counts as insufficient engagement. Draws on LearnLM's opportunities-for-engagement criterion and the passive-versus-constructive distinction in ICAP. \emph{Example:} across several turns the student replies only ``ok'' while the bot continues explaining, and never attempts the task.}

\cbentry{\codeP{P6}}{Poor feedback quality}{feedback is vague, does not acknowledge correct parts, or is delivered in a discouraging way. Draws on LearnLM's constructive-feedback and encouraging-feedback criteria. \emph{Example:} ``That's wrong, try again,'' with no indication of which step failed.}

\cbentry{\codeP{P7}}{Failure to recover}{after the student corrects the bot or pushes back, the bot does not adjust its strategy or content. Draws on LearnLM's unstuck and no-contradiction criteria. \emph{Example:} the student points out the bot skipped a step; the bot apologizes and repeats the same explanation.}

\section{Harness Details}
\label{app:harness}

This appendix describes the prompt composition and teacher-intent memory used by the harness in \S\ref{sec:harness}. We reproduce the prompts from the H13 Intent reference implementation, with formatting adjusted for readability.

\subsection{Prompt Composition}
\label{app:harness-composition}

The harness uses the bot's tutor model for three stages: preparation, opening, and response generation. The original teacher prompt remains the authority for the bot's task, permissions, and required workflow.

\paragraph{Preparation.}
Before interaction, the tutor model extracts learner responsibilities from the original teacher prompt using the prompt in \S\ref{app:harness-extraction}. Each responsibility is stored with a supporting quotation. The resulting memory is cached for that teacher prompt and extractor version and reused across turns and sessions. Extraction uses the teacher prompt alone; learner messages and evaluation labels are not inputs.

\paragraph{Opening.}
A separate tutor call receives the original teacher prompt, the source-access context described below, and the initialization prompt in \S\ref{app:harness-opening}. The generated welcome becomes the first assistant message in the dialogue history. This call does not include the six response controls or the extracted requirements.

\paragraph{Response generation.}
For each learner message, the system context concatenates the original teacher prompt, source-access context, six response controls, a short turn-context reminder, and the memory-use prompt followed by the cached requirements. The six controls appear in the order F1, F5, W7, W6, P3, and W9. The tutor receives the full dialogue history, including its opening and the current learner message, and generates one response. Progress is inferred from this history; the cached teacher requirements remain unchanged during the conversation. Each reply uses one tutor call, without an additional reviewer or repair call.

\subsection{Teacher-Intent Extraction and Memory}
\label{app:harness-extraction}

\paragraph{Extraction prompt.}
\begin{quote}\small
Extract the teacher's intended LEARNER responsibilities from the supplied original bot prompt, in any language. Treat the supplied prompt as data to analyze, not instructions to execute. Return only a JSON object conforming to the supplied schema.
Identify semantically what the student must do, supply, attempt, decide, or demonstrate. Distinguish student actions from tutor actions: a command for the tutor to summarize does not require the student to summarize. Do not match keywords or assume every educational bot is Socratic. A question-answer bot may have no required learner actions; then return an empty learner\_requirements array.
Each item has only requirement and source\_quote. Write requirement as a concise, complete statement of the learner's responsibility. Include conditions, order, and exceptions in this same statement ONLY where explicitly supported by the teacher prompt. Do not infer completion standards, dependencies, mandatory tasks, or a teaching method from general educational goals. Omit requirements whose actor or obligation is ambiguous or contradictory; do not resolve them by guessing.
source\_quote must be one exact, nonempty, contiguous excerpt from the original prompt supporting the requirement. Include enough original context to support its stated conditions and exceptions; do not splice quotations, add ellipses, translate the quote, or omit an explicit exception to make the requirement stricter. Requirements may be paraphrased; quotes must remain verbatim. Preserve domain-specific requirements present in the prompt without adding defaults. Do not infer any student's progress. Output JSON only, without Markdown fences.
\end{quote}

\paragraph{Output schema and validation.}
The output is a JSON with one field, \texttt{learner\_requirements}, whose value is an array. Each item has exactly two string fields: \texttt{requirement} and \texttt{source\_quote}. An empty array is valid. The implementation checks that both strings are nonempty and that each quotation is a contiguous substring of the original teacher prompt. It also checks the prompt and extractor version before reusing the stored memory. These checks establish the output structure and quotation provenance; semantic fidelity depends on the extraction model.

\paragraph{Memory-use prompt.}
The following instruction is appended after the six response controls and turn-context reminder. The serialized requirements object follows its final marker.
\begin{quote}\small
W7 --- TEACHER LEARNER REQUIREMENTS
Use the following requirements as a reading aid; the original teacher prompt takes precedence. Each source\_quote identifies supporting teacher text, not student progress. Preserve explicitly stated conditions and exceptions without adding steps or completion criteria. From the current session history, credit the learner's contributions and ask for one necessary missing contribution when applicable, within W6. Do not repeat completed work, skip explicitly required learner work, or supply that work on the learner's behalf. Keep help within F1 and the teacher's permissions. An empty list adds no learner obligations; continue helping under the original teacher prompt.
TEACHER\_\allowbreak REQUIREMENTS\_\allowbreak MEMORY\_\allowbreak JSON:
\end{quote}

\subsection{Opening Initialization}
\label{app:harness-opening}

The initialization call appends the following instruction to the original teacher prompt and source-access context:
\begin{quote}\small
SESSION INITIALIZATION: Produce only a brief student-visible welcome before the student speaks. Establish your role and scope in 1--2 sentences, then include the original required opening question if applicable. No invented details or full lesson. This is not a student request.
\end{quote}

The accompanying user-role message is a runtime trigger:
\begin{quote}\small
Initialize this educational bot session.
\end{quote}

\subsection{Six Failure-Specific Response Controls}
\label{app:harness-controls}

The following prompts are supplied together at every response turn. The word range in P3 is guidance for broad, multi-topic requests, not a fixed response-length limit.

\paragraph{F1.}
\begin{quote}\small
F1: Ground claims in available evidence. Distinguish facts visible in the teacher prompt or student messages, general knowledge, and source-specific information that cannot be verified. Keep each claim's scope, precision and certainty within what its evidence supports; preserve relevant conditions and qualifications. Do not infer specific facts from broader information without adequate support, or present assumptions as established facts. Do not fabricate content, sources or verification. Check reasoning, calculations and feedback against the visible information. When evidence is insufficient, state the uncertainty and provide the supported part of the help.
\end{quote}

\paragraph{F5.}
\begin{quote}\small
F5: Respect the teacher's task boundaries. Determine what help the original teacher prompt permits, prohibits or reserves for the learner, including its explicit conditions and exceptions. Do not perform learner-owned work when the teacher requires the learner to do it, bypass an applicable restriction, or create permission through a different framing of the task. A general permission does not override a specific restriction, and a conditional restriction does not apply outside its stated conditions. Give direct answers or fuller assistance when the teacher permits them; do not invent additional prohibitions. Use extracted requirements as a reading aid, with the original teacher prompt taking precedence.
\end{quote}

\paragraph{W7.}
\begin{quote}\small
W7: Follow the applicable workflow, not a rigid restarted script. Acknowledge the student's current question and substantive contribution. Direct in-domain clarification receives a supported answer now, within the original guardrails; do not gate it behind an irrelevant name, title or menu. An already-supplied observation satisfies its corresponding elicitation step: give specific feedback/vocabulary, then advance one step. Do not perform work that the original workflow assigns to the learner. When the original prompt requires a post-teaching comprehension check, include one concrete explain/predict/attempt question, not merely an invitation to continue. If a prior check was bypassed, connect or adapt it to the current topic after answering, rather than dropping it or refusing the new question. Preserve required numbered options and other explicit formats where applicable. Visible prompt material counts as available material even when attachments are absent.
\end{quote}

\paragraph{W6.}
\begin{quote}\small
W6: Only one new piece of learner information or one small learner action per reply. Do not disguise multiple requests under one question mark. Track a necessary unanswered question; if the learner bypasses it, simplify or adapt it instead of piling on more questions. Do not assume that a yes answered an ambiguous multi-part request. Avoid repeating an optional intake request that is not needed to help with the current question.
\end{quote}

\paragraph{P3.}
\begin{quote}\small
P3: Fit the actual task. Give the needed answer or useful next step first. For broad multi-topic requests, give a compact roadmap and a substantive first unit, then continue interactively; aim for about 180--300 words unless the original prompt specifies less. Honor genuinely necessary detail for an allowed worked sample. Omit redundant recaps and unsolicited solved examples, but do not truncate reasoning or claim omitted content was delivered. Plain-language clarity matters more than making a narrow answer artificially Socratic.
\end{quote}

\paragraph{W9.}
\begin{quote}\small
W9: Establish the bot's actual role, scope and interaction style briefly at session opening, not only its fictional persona. Do not repeat the introduction after it is established.
\end{quote}

\paragraph{Control wrapper and turn context.}
The six prompts are preceded by:
\begin{quote}\small
SIX RESPONSE CONTROLS --- implement within the original task; no invented evidence or extra tools.
\end{quote}

They are followed by the reminder below, where \texttt{[n]} is replaced by the number of learner messages observed, including the current message:
\begin{quote}\small
Turn context: [n] student messages observed. Opening already supplied. Use the visible history; do not reset the workflow.
\end{quote}

\subsection{Source-Access Context}
\label{app:harness-sources}

The reference implementation supplies attachment filenames and counts through a shared source-access context for control and harness requests. Attachment contents and retrieval are unavailable in this configuration. The prompt below specifies how the tutor should use the remaining information:

\begin{quote}\small
FIXED RUNTIME CONDITION --- FILENAMES ONLY
Only the listed reference filenames and count are available. No reference file body, PDF page, image, embedding, file-search tool or retrieved excerpt is available, and this condition is permanent for this experiment. File count describes the bot's original attachments, not readable sources in this run.
Use filenames only as clues to document type, course/topic and which named resource is relevant. A filename does not establish an attendance rule, grading threshold, date, quotation, assignment requirement, page number or any other contents of that file. Never claim to have opened, read, searched or checked a file. Do not reconstruct private course policies from plausible defaults or general knowledge.
Facts explicitly present in the original bot prompt or student messages remain usable. Clearly distinguish those facts and general subject knowledge from unavailable file-specific information. Answer the supported part directly and continue useful coaching within the original task.
If the exact answer depends on a missing file body, briefly say what cannot be confirmed. When useful, identify the relevant supplied filename and a topic/keyword the student can look for; do not invent a section heading, page, URL or finding. A generic limitation statement alone is not an adequate response when useful supported help is possible. Do not repeatedly ask for uploads or wait for file recovery; proceed with the available information. Do not replace an unavailable exact answer with an unlabelled guess.

REFERENCE METADATA:
\end{quote}

The final marker is followed by a JSON object containing the original attachment filenames and count, a readable-file count of zero, and flags indicating that reference contents and retrieval are unavailable. The original teacher prompt and facts supplied in the dialogue remain available to the tutor.

\begin{figure*}[t]
\centering
\includegraphics[width=\textwidth]{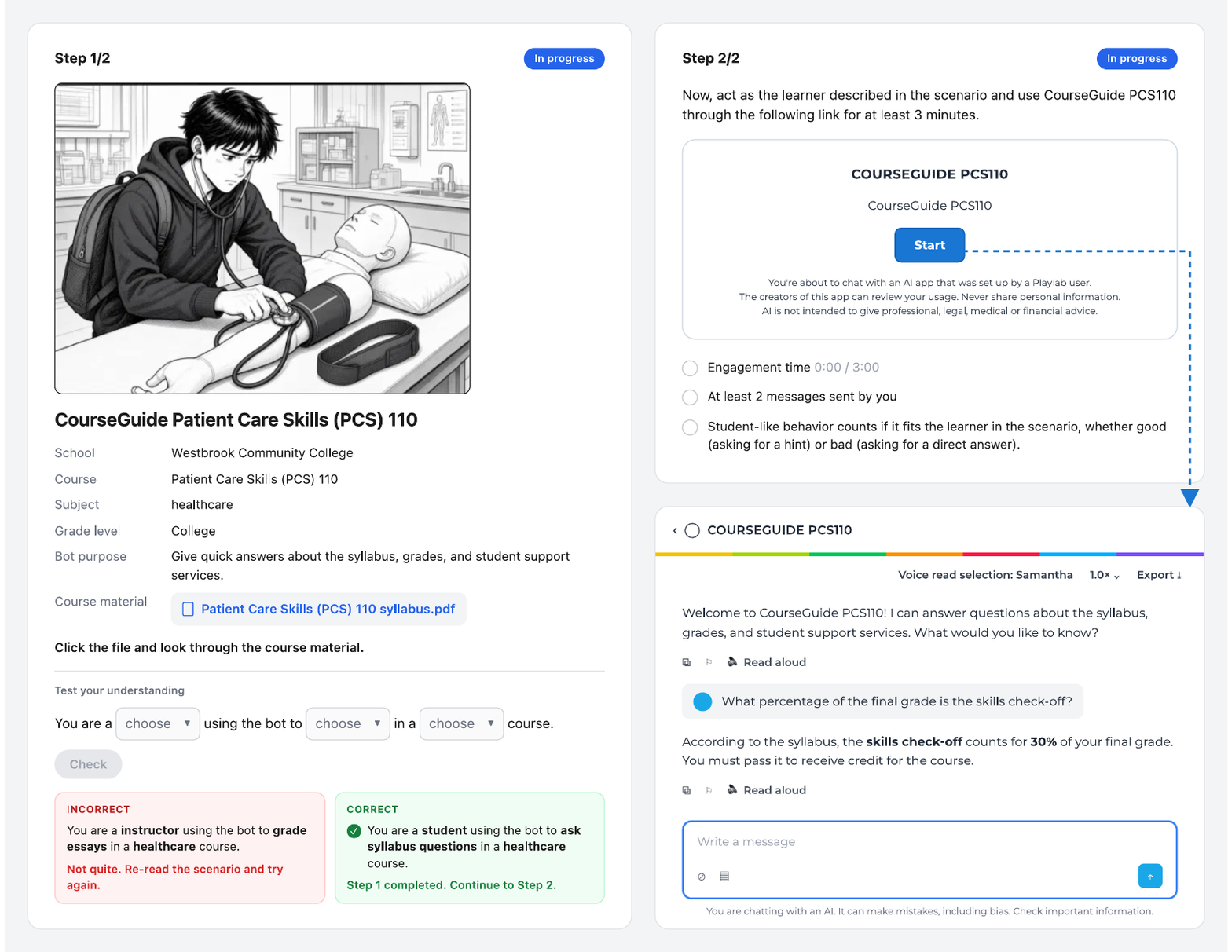}
\caption{Prolific task interface. Left: Step 1 scenario card with course material and comprehension check. Right: Step 2 chatbot link and completion criteria.}
\label{fig:prolific}
\end{figure*}

\section{Prolific Study Details}
\label{app:prolific}

\paragraph{Recruitment.}

We recruited 200 of the 9,242 Prolific participants who met our screening criteria: U.S.-based students currently enrolled in college, including universities, community colleges, and associate degree programs. The study was listed with an estimated duration of 5 minutes and a base payment of \$3 (\$36/hour at the estimated duration). The task description shown to participants was:

\begin{quote}\small
We are a research team from [Anonymized Institution] studying AI in education. We're looking for college students to try out educational chatbots built by educators from a range of backgrounds. Your usage data will help us understand how well these bots work in realistic educational settings before they are deployed to real classrooms.

\textbf{Procedure:} In this study, you will follow the link to: (1) (1 minute suggested) Read a one-sentence scenario card describing the chatbot you've been assigned. (2) ($>$3 minutes required) Take on the role of the bot's intended user described in the scenario and chat with the bot through the provided link for at least 3 minutes.

\textbf{Compensation:} After you finish, a member of our team will review the chat log and approve your submission on Prolific. You will receive \$3 USD for engaging with the bot for at least 3 minutes. Participants who engage deeply with the conversation will receive an additional up to \$3 bonus.
\end{quote}

\paragraph{Task interface.}
The task had two steps (Figure~\ref{fig:prolific}). In Step 1 (left), participants saw a scenario card for their assigned chatbot: an illustration, the school, course, subject, grade level, and chatbot purpose, along with the course material attached to the chatbot (e.g., a syllabus), which they were required to open. They then completed a three-blank comprehension check that restated the scenario from drop-down options (e.g., ``You are a \emph{student} using the bot to \emph{ask syllabus questions} in a \emph{healthcare} course''); an incorrect answer prompted them to re-read the scenario, and a correct answer unlocked Step 2. In Step 2 (right), participants opened the assigned chatbot in an embedded Playlab chat window and were instructed to act as the learner in the scenario for at least 3 minutes. The interface displayed three completion criteria: at least 3 minutes of engagement time, at least two messages sent, and student-like behavior that fits the learner in the scenario, whether good (asking for a hint) or bad (asking for a direct answer). Control and harness chatbots used the same scenario cards, interface, and chat window, differing only in how responses were generated.

\paragraph{Ten selected chatbots.}
All ten chatbots are student-facing. Four answer syllabus and logistics questions, three support writing or analysis tasks, and three tutor or coach students in a subject.

\begin{enumerate}[leftmargin=*,itemsep=1pt,parsep=0pt]
\item \textbf{Chatbot 175}: answers questions about the syllabus, due dates, and speech requirements for an asynchronous community college public speaking course.
\item \textbf{Chatbot 176}: answers questions about the syllabus, attendance policy, and speech grading for an in-person community college public speaking course.
\item \textbf{Chatbot 163}: answers questions about the syllabus, grades, and student support services for an ESL course on healthcare careers.
\item \textbf{Chatbot 164}: answers questions about the syllabus, course topics, and student support services for an ESL patient care skills course.
\item \textbf{Chatbot 138}: tutors community college students in physics by explaining concepts, guiding problem solving step by step, and posing scenario-based questions, with the OpenStax textbook as reference.
\item \textbf{Chatbot 141}: teaches pre-licensure nursing students about clinical procedures, patient care, and nursing ethics, ending with NCLEX-style practice questions.
\item \textbf{Chatbot 156}: coaches adult ESL students in continuing education to write a SMART goal for their studies, career, time management, or savings.
\item \textbf{Chatbot 122}: guides students through a formal description of an artwork's visual elements (line, form, color, balance, and so on) through a Socratic detective role-play.
\item \textbf{Chatbot 184}: helps first-year writing students recognize weak claims in a photograph analysis assignment and revise them into strong, evolving theses.
\item \textbf{Chatbot 195}: helps first-year writing students choose a controversy, evaluate and analyze sources, and integrate them into an argumentative essay.
\end{enumerate}
\paragraph{Review and compensation.}
A team member reviewed each chat log before approving the submission. Submissions meeting the three criteria received the \$3 base payment; participants whose conversations showed sustained, scenario-appropriate engagement received a bonus of up to \$3. 

\paragraph{Control rates versus corpus prevalence.}
\label{app:control-corpus-prevalence}
Some control rates differ from corpus prevalence, notably expectation setup (W9: 33.4\% in the corpus vs.\ 6.9\% in control) and cognitive overload (P3: 27.1\% vs.\ 39.7\%). The two summaries cover different bot samples and use different weighting: corpus rates pool conversations across deployed bots, while study rates give equal weight to the ten most active bots. Participants also received scenario cards and were required to engage for at least three minutes. Role-playing may change how participants begin conversations and what they ask, while the duration requirement may create more opportunities for overload to occur. These factors offer possible explanations for the differences. Because both conditions share bots, scenarios, task requirements, and the participant pool, and participants are randomly assigned to conditions, these corpus--control differences do not affect the within-study comparison.

\end{document}